\pdfoutput=1

\documentclass[11pt]{article}

\usepackage[final]{acl}

\usepackage{times}
\usepackage{latexsym}

\usepackage[T1]{fontenc}

\usepackage[utf8]{inputenc}

\usepackage{microtype}

\usepackage{inconsolata}

\usepackage{graphicx}

\usepackage{bbm}
\usepackage{enumitem}
\usepackage{tcolorbox}
\usepackage{hyperref}
\usepackage{amsmath}
\usepackage{amssymb}
\usepackage{booktabs}
\usepackage{multirow}
\usepackage{makecell}
\usepackage[capitalise]{cleveref}

\title{Expanding the Boundaries of Vision Prior Knowledge \\ in Multi-modal Large Language Models}

\author{
  \textbf{Qiao Liang}${}^{1,2}$\textsuperscript{\textnormal{*}},
  \textbf{Yanjiang Liu}${}^{1,2}$\textsuperscript{\textnormal{*}}, 
  \textbf{Weixiang Zhou}${}^{2}$,
  \textbf{Ben He}${}^{1,2}$,
  \textbf{Yaojie Lu}${}^{2}$, 
  \textbf{Hongyu Lin}${}^{2}$,\\
  \textbf{Jia Zheng${}^{2}$},
  \textbf{Xianpei Han${}^{2}$},
  \textbf{Le Sun${}^{2}$},
  \textbf{Yingfei Sun${}^{1}$}
  \\
  ${}^{1}$University of Chinese Academy of Sciences \\
  ${}^{2}$Chinese Information Processing Laboratory, Institute of Software, Chinese Academy of Sciences\\
 \texttt{\{liangqiao2022,liuyanjiang2021,weixiang,luyaojie,hongyu,zhengjia\}@iscas.ac.cn}\\
  \texttt{\{benhe,yfsun\}@ucas.ac.cn}, \texttt{\{xianpei,sunle\}@iscas.ac.cn}
}

\begin{document}
\maketitle
\begin{abstract}

Does the prior knowledge of the vision encoder constrain the capability boundary of Multi-modal Large Language Models (MLLMs)? While most existing research treats MLLMs as unified systems optimized through end-to-end training, the impact of vision encoder's prior knowledge is seldom investigated. In this work, we introduce a novel metric, $Rank_e$, to quantify the effect of prior knowledge of the vision encoder on MLLM performance. Our analysis reveals a positive correlation between prior knowledge and MLLM performance. Moreover, we find that domain-specific fine-tuning using solely end-to-end visual question answering (VQA) data is insufficient, particularly for entities with low inherent visual prior knowledge. To address this issue, we propose VisPRE (Vision Prior Remediation), a two-stage training framework that explicitly incorporates prior knowledge at the vision encoder level. Experimental results demonstrate that augmenting vision encoder’s prior knowledge substantially boosts the visual understanding capabilities of MLLMs, offering a novel and effective strategy for improving performance, especially in scenarios involving uncommon visual entities.

\renewcommand{\thefootnote}{\fnsymbol{footnote}}

\footnotetext[1]{Equal contribution.}
\end{abstract}
\section{Introduction}
\label{sec:intro}

Multi-modal Large Language Models have emerged as a rapidly growing area of research. Combining the powerful capabilities of Large Language Models with the ability to process visual input, MLLMs excel in tasks such as image understanding, VQA~ \cite{agrawal2016vqavisualquestionanswering}, image captioning, and visual instruction following. The development of models such as GPT-4o \cite{4oapi}, GPT-4V \cite{4v}, and Claude-3.5 \cite{claude35} have demonstrated remarkable proficiency in advanced multi-modal understanding. Besides, open-source models like LLaVA \cite{liu2024visual, liu2024improvedbaselinesvisualinstruction, li2024llavaonevisioneasyvisualtask} series, Qwen2-VL \cite{wang2024qwen2vlenhancingvisionlanguagemodels}, and InternVL2 \cite{chen2024internvl, chen2024far} are making significant strides, bridging the gap in the field. 

\begin{figure}[t]
  \centering
  \includegraphics[width=\linewidth]{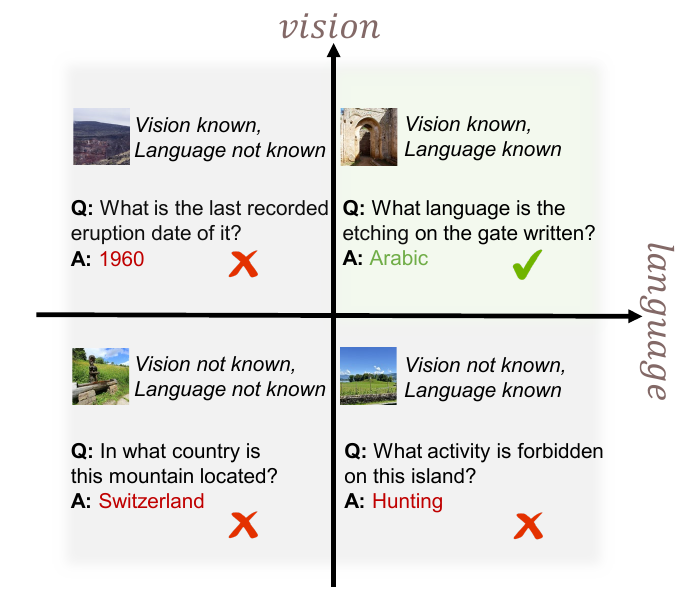}
  
  \caption{\textbf{Knowledge quadrants of a MLLM.} ``Vision known'' indicates that the vision encoder recognises the entity in the image, while ``Language known'' indicates that the language model possesses relevant information about the entity. Only when both vision and language are ``known'' can the MLLM achieve accurate comprehension and response generation. }
  \label{fig:shouye}
\end{figure}

\begin{figure*}[t]
  \centering
  \includegraphics[width=0.8\linewidth]{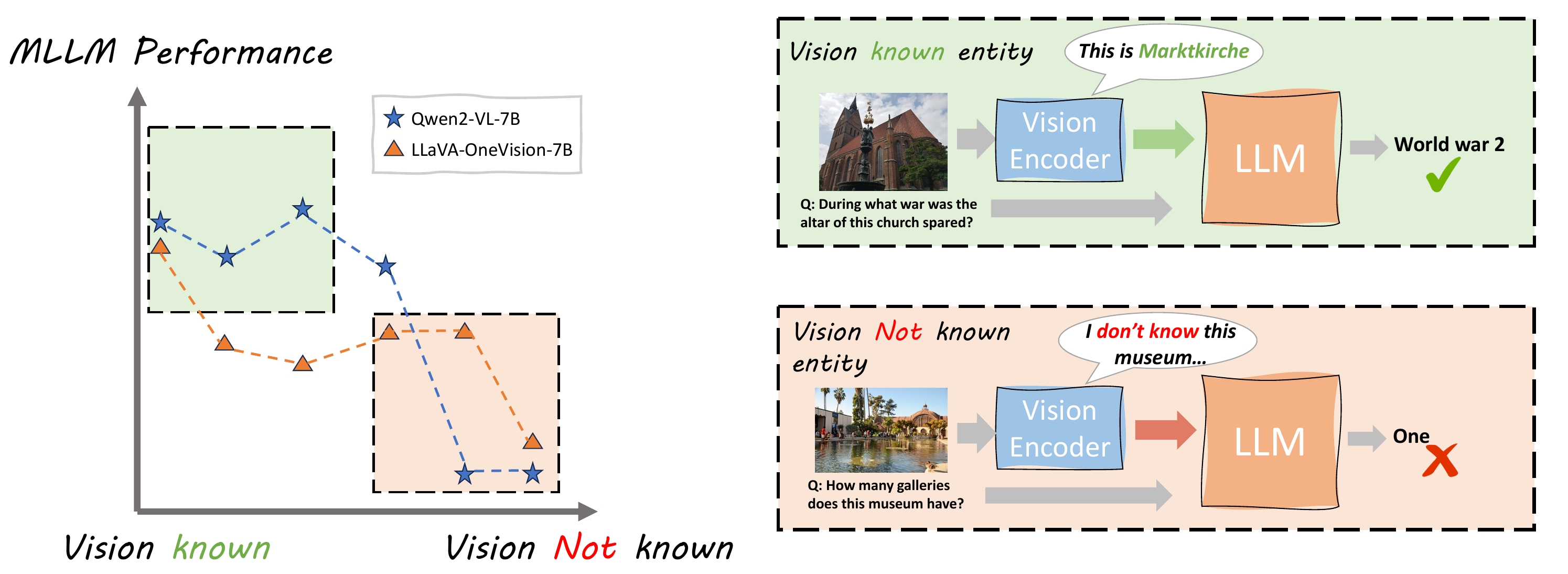}
  
  \caption{Left: \textbf{Current MLLM performance vs. vision prior knowledge.} Current MLLMs demonstrate positive correlation between vision prior knowledge and overall performance. Right: \textbf{``Vision Known'' and ``Vision Not Known'' Entities.} (1) For ``vision known entities'', the vision encoder contains sufficient prior knowledge, enabling MLLM answers correctly; (2) For ``vision not known entities'', insufficient visual knowledge leads to MLLM failure. We propose the $Rank_e$ metric to quantify vision encoder's prior knowledge about specific entities, along with a two-stage training framework to enhance encoder knowledge, expanding MLLM's performance boundaries.}
  \label{fig:concept}
\end{figure*}

A pivotal challenge in advancing MLLMs is forging a seamless and robust alignment between vision and language. One effective approach involves integrating an off-the-shelf external vision encoder with a language model using a modality conversion module~ \cite{alayrac2022flamingo, li2023blip2bootstrappinglanguageimagepretraining, li2023lmeyeinteractiveperceptionnetwork,zhu2023minigpt4enhancingvisionlanguageunderstanding,dai2023instructblipgeneralpurposevisionlanguagemodels, bai2023qwen, liu2024visual, li-etal-2022-mplug},  which we refer to as the modular approach. Compared to the monolithic multi-modal approach \cite{team2024chameleon, luo2024mono, fuyu-8b, zhan-etal-2024-anygpt}, which is built from scratch using multi-modal data, the modular approach is more data-efficient and achieves comparable performance. Despite these advantages, the modular approach still faces challenges, as the vision and language components are trained separately from distinct data distributions, leading to an inherent misalignment in their knowledge.
To illustrate the importance of knowledge alignment, we present a knowledge quadrant diagram in \cref{fig:shouye}, with the horizontal axis representing the knowledge held by the language model and the vertical axis representing the knowledge held by the vision encoder. Only when both components possess necessary knowledge (in the ``Vision known \& Language known'' quadrant) can the multi-modal model accurately handle complex cross-modal tasks~ \cite{li2023multimodalcontextreasoningapproach,cheng2024aiassistantsknowdont}. Misalignment in knowledge from either the vision or language side introduces limitations to the model’s capabilities, making it essential to bridge this gap to enhance the performance of multi-modal models. Many existing studies focus on addressing knowledge misalignment from the language perspective, expanding from ``Vision known \& Language not known'' to ``Vision known \& Language known''.  Some studies \citep{ caffagni2024wiki,jiang2024understandingrolellmsmultimodal} enhance language model knowledge with external documents related to images, while CVLM \cite{li2024cognitive} trains a ``Visual Knowledge Aligner'' module to enrich text-based knowledge associated with images. However, as a crucial component of MLLM~ \cite{collins2014knowledge}, the vision encoder also possesses varying prior knowledge about the real world, such as entities, textures, and causality~ \cite{PINKER19841, CAVANAGH20111538}. But the impact of this vision prior knowledge on MLLM capabilities remains unexplored, leading to a natural question: \textbf{How does vision prior knowledge affect MLLM's capability?} In this paper, we attempt to answer this question by investigating the following research questions:

\begin{itemize}[leftmargin=*]
\item \textbf{Q1:} How to measure prior knowledge in vision encoders?
\item \textbf{Q2:} Does vision prior knowledge constrain MLLM?
\item \textbf{Q3:} How to transcend vision prior knowledge limits?
\end{itemize}


To address these questions, we introduce $Rank_e$ to quantify the vision encoder's prior knowledge. Through experiments with various model combinations, we reveal a positive correlation between MLLM performance and visual prior knowledge. \cref{fig:concept} (left) demonstrates the relationship between current MLLM performance and vision prior. Furthermore, we find that direct fine-tuning with end-to-end VQA data is insufficient for improving MLLM performance on low prior entities. \cref{fig:concept} (right) illustrates the knowledge misalignment on low prior entities. To overcome this limitation, we propose a two-stage training framework that injects vision prior knowledge into the vision encoder, resulting in significant improvements in MLLM performance. In summary, our main contributions are:



\begin{itemize}
    \item We introduce the $Rank_e$ metric to quantify a vision encoder's prior knowledge, revealing a positive correlation between MLLM performance and the encoder's embedded visual knowledge.
    
    \item Our analysis shows that domain-specific finetuning with only end-to-end VQA data proves insufficient, particularly for entities with low vision prior knowledge.
    
    \item We propose a two-stage training framework \textbf{VisPRE} (\textbf{Vis}ion \textbf{P}rior \textbf{Re}mediation) that injects prior knowledge at the vision encoder level, significantly enhancing MLLM performance, especially for entities with low vision prior knowledge.
\end{itemize}

\section{Vision Prior Measurement}
\label{sec:formulation}

Vision encoders are typically trained on extremely large-scale data (from 400 million to 10 billion samples \cite{NEURIPS2024_9ee3a664}), often with undisclosed data (e.g., OpenAI CLIP \cite{radford2021learning}), making direct evaluation of vision priors from training data infeasible. Therefore, to answer \textbf{Q1}, we shift our focus to evaluating observable behavioral evidence - specifically, how effectively these encoders recognize visual entities. We thus propose the $Rank_e$ metric, which quantifies an encoder's vision prior knowledge for a given entity $e$.

In this section, we begin by describing the modality alignment process in modular MLLMs, then formulating the definition of vision prior knowledge. Finally, we introduce the $Rank_e$ metric to quantify this knowledge.

Modular MLLMs establish cross-modal understanding through an alignment process that maps visual information to textual representations. Formally, given an input text prompt $T_A$ and target image $I_B$, where $\mathcal{F}$ represents the MLLM's internal representation function that maps inputs to hidden states, the alignment process can be described as: 


\begin{equation}
    \begin{split}
        \mathcal{F}(T_A, I_B) \xrightarrow{\text{align}} \mathcal{F}(T_A, \hat{T}_B) \\
        \text{where} \quad \hat{T}_B \sim P(T|I_B)
    \end{split}
\end{equation}

Here, $\hat{T}_B$ represents the generated text that preserves the semantic content of $I_B$.
Building upon the Platonic representation hypothesis \cite{huh2024platonic}, we posit that cross-modal alignment occurs through a shared latent space $\mathcal{Z}$. This allows us to decompose the \(P(T|I_B)\) as:

\begin{equation}
    P(T|I_B) = \sum_{z\in\mathcal{Z}} \underbrace{P_{\text{vision}}(z|I_B)}_{\text{Vision prior}} \cdot P_{\text{align}}(T|z, I_B)
\end{equation}

The latent representation $z$ serves as an intermediary that connects the visual and textual domains. While \( P_{\text{align}}(T|z, I_B) \) reflects the MLLM's ability to convert latent representation $z$ into textual output $T$, \( P_{\text{vision}}(z|I_B) \) represents the vision encoder's capability to transform image $I_B$ into an appropriate latent representation. \( P_{\text{vision}}(z|I_B) \) constitutes what we define as vision prior knowledge—the encoder's pre-existing understanding of visual entities encoded in its parameters.

\begin{figure}[t]
  \centering
  \includegraphics[width=\linewidth]{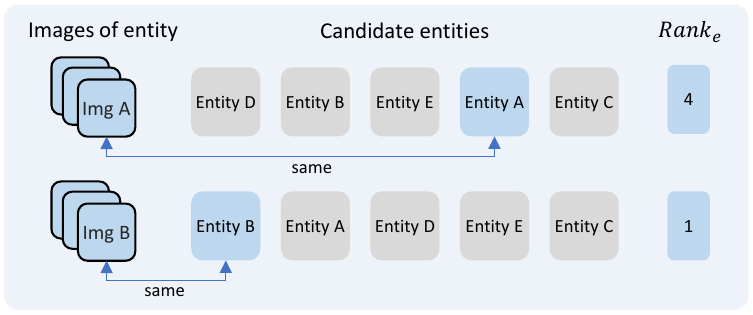}
  
  \caption{\textbf{Illustration of metric \text{Rank}$_e$.} For a target entity \( e \), we compute cross-modal similarity scores between its vision representations (extracted by vision encoder) and text representations of all candidate entities (extracted by corresponding text encoder). The rank of entity \( e \) among these candidates defines its $Rank_e$. In this example, while Image A depicts Entity A, entity A achieves 4th-highest similarity score, resulting in $Rank_e$ = 4.}
  \label{fig:rank}
\end{figure}

To quantify the inherent vision prior \( P_{\text{vision}}(z|I_B) \), we discretize the continuous latent space \( \mathcal{Z} \) into a set of entity-specific latent representations. For a given image \( I_B \), we approximate \( P(z|I_B) \) by evaluating the probability that the vision encoder correctly identifies an entity within \( I_B \).  To achieve this, we propose the \(Rank_e\) metric, which measures how well the encoder identifies a target entity \( e \) from visual inputs, thereby evaluating the vision encoder's inherent prior knowledge.  As shown in \cref{fig:rank}, for an entity \( e \), \(Rank_e\) is computed as follows:
\begin{itemize}[leftmargin=*]
    \item \textbf{Similarity scoring:} For an image $I_e$ containing entity \(e\), compute the image-text similarity score $s_{j} = \phi(I_e, T_j)$ using the vision encoder and its corresponding text encoder, where $\{T_1, ..., T_n\}$ are textual descriptions of $n$ candidate entities.
    \item \textbf{Ranking:} Rank the entities in descending order based on their similarity scores $\{s_{j}\}_{j=1}^n$, and record the position of the target entity $e$ as \(Rank_e\). If  multiple images  \(\{I_e^{(1)}, ..., I_e^{(m)}\}\) are available for single entity \(e\), compute \(Rank_e\) for each image separately and take the average:
    \begin{equation}
        \text{Rank}_e = \frac{1}{m} \sum_{i=1}^{m} rank\big(\phi(I_e^{(i)}, T_e)\big).
    \end{equation}
    where \(rank(\phi (I_e, T_e))\) denotes the position of \(\phi (I_e, T_e))\) in ordered \(\{s_j\}_{j=1}^n\). Lower $Rank_e$ values indicate stronger visual prior knowledge, with optimal performance when $\text{Rank}_e = 1$.
\end{itemize} 

\section{Experiments}
\label{sec:experienment}

In this section, we explore the three proposed research questions. In Section \ref{sec:3-1}, we describe the overall experimental setup. In Section \ref{sec:3-2}, we verify the relationship between MLLM and the prior knowledge of its vision encoder. From Section \ref{sec:3-3} to Section \ref{sec:3-4}, we show the insufficiency of end-to-end fine-tuning and propose a training framework to transcend vision prior knowledge limits.

\subsection{Experiment Setting}
\label{sec:3-1}
\textbf{Models.} To systematically examine the impact of vision encoder’s prior knowledge on MLLM performance across different vision encoders and base LLM combinations, we train nine MLLMs from scratch based on an encoder-projector-LLM architecture. For the vision encoder, we use widely adopted encoders in MLLMs, including OpenAI ViT-L-14 \cite{radford2021learning}, SigLIP ViT-SO-14 \cite{zhai2023sigmoid}, and DFN ViT-H-14 \cite{fang2023data}. For base LLM, we select the LLaVA-1.5 language model, Vicuna-7B-v1.5 \cite{vicuna2023}, and recent open-source models, Llama-3.1-Instruct-7B \cite{dubey2024llama} and Qwen-2.5-Instruct-7B \cite{qwen2.5}. 

\textbf{Datasets.} To evaluate MLLMs under different vision priors, we require a VQA dataset that meets two conditions: (1) it provides entity annotations covering a wide range of prior knowledge—from extremely rare to very common entities; (2) it includes entity-centric visual questions and answers for MLLM performance assessment. Here, rare entities refer to those that appear infrequently or not at all in the vision encoder's training data, making them difficult for the vision encoder to recognize accurately. The Encyclopedia-VQA \cite{mensink2023encyclopedic} dataset fulfills both requirements. With extensive entity annotations covering up to 16.7k entity categories, it captures both common and rare entities and poses a hard challenge for MLLMs with its knowledge-based VQA questions.

\textbf{Training.} We conducted training on a 8×A800 GPUs. Initially, we pre-trained the model on the LLaVA \cite{liu2024visual} dataset to develop an MLP projector aligned with selected vision encoder. For fine-tuning phase, we sampled 10\% of the LLaVA instruction tuning dataset and integrated it with additional fine-tuning data to optimize computational efficiency while maintaining performance quality.

\textbf{Metrics and Evaluation.} We use Llama-3.1-70B \cite{dubey2024llama} to judge model responses, denoted as a function \( g(\cdot) \) that takes the question, entity, ground truth answer, and model output as input, returning \textit{true} if the answer is correct. Using this, we define entity accuracy \( \text{Acc}_{e} \) for each entity \( e \) as the fraction of correct responses among all related questions:

\begin{equation}
\text{Acc}_{e} = \frac{1}{N_e} \sum_{i=1}^{N_e} \mathbbm{1}\left[ g(y_i, \hat{y}_i) = \text{true} \right]
\end{equation}

where \( N_e \) is the number of questions for entity \( e \), \( y_i \) is the ground truth answer and other question information, and \( \hat{y}_i \) is the model's output. The overall dataset accuracy \( \text{Acc}_{\text{macro}} \) is calculated as the macro-average of all entity accuracies. Details of the evaluation configurations are in Appendix \ref{app:b}.

\begin{figure*}[t]
  \centering
  \includegraphics[width=1\linewidth]{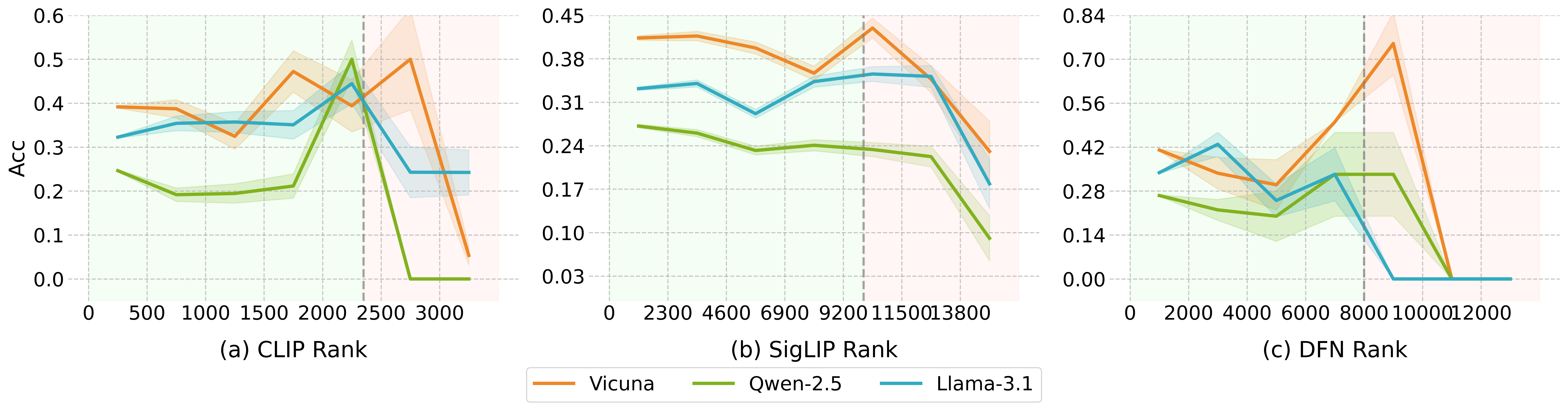}
  \caption{\textbf{MLLM Performance distribution across different Rank$_e$ intervals.} Performance of all MLLMs decreases as $Rank_e$ increases across three encoder configurations. The Vicuna-CLIP model shows an 87\% performance drop from $0<Rank_e<500$ to $Rank_e>3000$, indicating correlation between performance and vision prior knowledge. This relationship is non-linear with a critical threshold. We marked this threshold by a vertical line in the figure—green on the left indicating sufficient prior knowledge for reasoning, and red on the right showing insufficient knowledge causing sharp performance decline.}
  \vspace{-2mm}
  \label{fig:all_rank}
\end{figure*}


\subsection{Vision Prior Constrains MLLM Performance}
\label{sec:3-2}
To investigate \textbf{Q2:} ``Does vision prior knowledge constrain MLLM?'', we first categorize entities into two types: those ``vision encoder knows'' and those ``vision encoder doesn't know'' then observe MLLM performance across both categories. Through our proposed $Rank_e$ metric, we measure the vision encoder's knowledge of entities in Encyclopedia-VQA, where a lower $Rank_e$ indicates greater knowledge. For MLLM performance, we test accuracy in answering entity-related questions in Encyclopedia-VQA.

Our study aims to address knowledge misalignment where MLLM capabilities are limited by the vision encoder. Therefore, we retain only cases where the LLM component possesses adequate entity knowledge, regardless of the vision encoder's knowledge. Specifically, we prompt the MLLM with ``This is \{entity name\}'' rather than the actual image; if the MLLM answers correctly, we retain this case. Additionally, we discovered a number of cases where MLLMs provide correct answer without image description or actual image. We attribute this to the MLLM's dependency on question format \cite{jiang2024understandingrolellmsmultimodal}. We eliminated this subset from our analysis. \cref{fig:all_rank} illustrates the relationship between MLLM accuracy and $Rank_e$

\textbf{Finding 1: MLLM performance correlates positively with vision prior knowledge.} As shown in \cref{fig:all_rank}, across all three encoder choices, MLLM performance consistently declines as entity $Rank_e$ increases. For the CLIP encoder, from the interval $0 < Rank_e < 500$ to $Rank_e > 3000$, Vicuna's performance drops by 87\%, Llama3.1's by 100\%, and Qwen-2.5's by 21\%. In SigLIP encoder experiments, overall performance declines by about 50\% across all three models from the leftmost to the rightmost interval, while for the DFN encoder, the decline reaches 100\%.

Notably, CLIP-Vicuna MLLM does not exhibit a significant performance decline until $Rank_e$ reaches 3000. The phenomenon is also observed in the SigLIP and DFN configurations. This threshold effect suggests that the positive correlation between vision prior knowledge and MLLM performance is not strictly linear, but rather exhibits a mutation beyond a critical point. We posit that this stems from the vision encoder holding a \textit{known} status for entities below a certain $Rank_e$ threshold, meaning it can still provide sufficient prior knowledge for the MLLM to answer entity-related questions. Once $Rank_e$ exceeds this threshold, the vision encoder no longer provides adequate prior knowledge, resulting in a sharp drop in MLLM performance. Considering that LLM part of MLLM possesses adequate knowledge about all entities here, it is the vision encoder of MLLM that constrains the overall performance on entities beyond the threshold.

\subsection{Shortcomings of End-to-end Finetuning}
\label{sec:3-3}

To investigate \textbf{Q3:} ``How to transcend vision prior knowledge limits?'', we implement a typical solution as our baseline—finetuning MLLMs on end-to-end domain-specific VQA data. Following established MLLM finetuning approaches \cite{liu2024visual, liu2024improvedbaselinesvisualinstruction}, we freeze the vision encoder parameters and only tune the LLM component. This setup enables the LLM parameters to compensate for limitations in vision prior knowledge.

\begin{table}[h]
\centering
\resizebox{0.45\textwidth}{!}{%
\begin{tabular}{ccccc}
\toprule
\multirow{2}{*}{Vision Encoder}   & \multirow{2}{*}{LLM} & \multicolumn{2}{c}{Number of (Q, A) pairs} & \multirow{2}{*}{\makecell{Number of\\entities}} \\ \cmidrule{3-4}
                                  &                      & \makecell{\hspace{1em}Train}               & \makecell{\hspace{1em}Test}                &                                                                               \\ \midrule
\multirow{3}{*}{OpenAI ViT-L-14}  & Vicuna-7B            & \makecell{\hspace{1em}1877}                & \makecell{\hspace{1em}531}                 & 90                                                                            \\
                                  & Llama3.1-8B          & \makecell{\hspace{1em}2305}                & \makecell{\hspace{1em}624}                 & 106                                                                           \\
                                  & Qwen2.5-7B           & \makecell{\hspace{1em}2345}                & \makecell{\hspace{1em}645}                 & 109                                                                           \\ \midrule
\multirow{3}{*}{SigLIP ViT-SO-14} & Vicuna-7B            & \makecell{\hspace{1em}2290}                & \makecell{\hspace{1em}615}                 & 106                                                                           \\
                                  & Llama3.1-8B          & \makecell{\hspace{1em}2669}                & \makecell{\hspace{1em}717}                 & 123                                                                           \\
                                  & Qwen2.5-7B           & \makecell{\hspace{1em}2614}                & \makecell{\hspace{1em}705}                 & 118                                                                           \\ \midrule
\multirow{3}{*}{DFN ViT-H-14}     & Vicuna-7B            & \makecell{\hspace{1em}1914}                & \makecell{\hspace{1em}531}                 & 90                                                                            \\
                                  & Llama3.1-8B          & \makecell{\hspace{1em}2339}                & \makecell{\hspace{1em}615}                 & 105                                                                           \\
                                  & Qwen2.5-7B           & \makecell{\hspace{1em}2291}                & \makecell{\hspace{1em}618}                 & 105

                                \\ \bottomrule
\end{tabular}


}
\caption{\textbf{Dataset Statistics.} We report the number of (question, answer) pairs for each dataset split across different encoder-language model combinations. Each corresponding train-test pair shares the same entities.
}
\vspace{-3mm}
\label{tab:data_statis}
\end{table}


We constructed our finetuning dataset from Encyclopedia-VQA. Following the method in Section \ref{sec:3-2}, we retained questions that MLLMs answered correctly when prompted with ``This is \{entity\_name\}'' instead of the actual image. After calculating $Rank_e$ across the dataset, we observed naturally different $Rank_e$ distributions across encoders. To balance the distribution of entities with varying levels of prior knowledge, we sampled entities to create more uniform rank distributions for validation. We then divided each subset into training and test sets containing the same entities but with different questions. Dataset statistics are presented in \cref{tab:data_statis}, with detailed construction methodology in Appendix \ref{app:a}.

Successful knowledge-based VQA requires three essential MLLM capabilities: (1) recognizing entities in images; (2) possessing relevant knowledge about these entities; and (3) utilizing this knowledge to answer questions. As the LLM component already contains adequate entity knowledge, MLLM performance can be enhanced through two approaches: (1) improving visual entity recognition and (2) optimizing knowledge utilization for question answering.

To explore these approaches, we develop two distinct types of finetuning data: (1) \textbf{Perception-tuning data}, where we transform original Encyclopedia-VQA questions into perception-focused queries such as \textit{What is this image about?} and (2) \textbf{Knowledge-tuning data}, which preserves the original questions from Encyclopedia-VQA. Detailed construction methodologies for both datasets are provided in Appendix \ref{app:a}.

\begin{figure}[t]
    \centering
    \begin{minipage}[b]{0.45\textwidth}
        \centering
        \includegraphics[width=\textwidth]{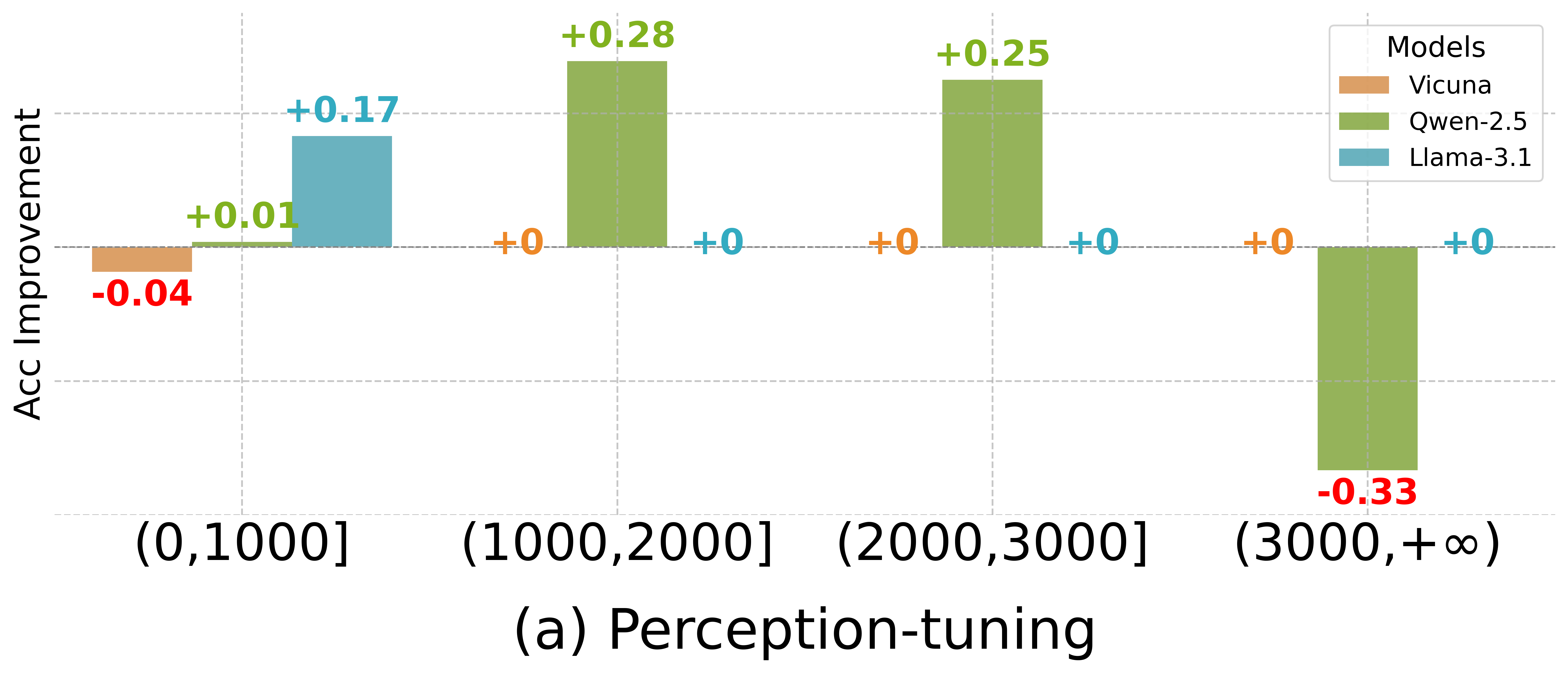}
    \end{minipage}
    \hspace{0.05\linewidth}
    \begin{minipage}[b]{0.45\textwidth}
        \centering
        \includegraphics[width=\textwidth]{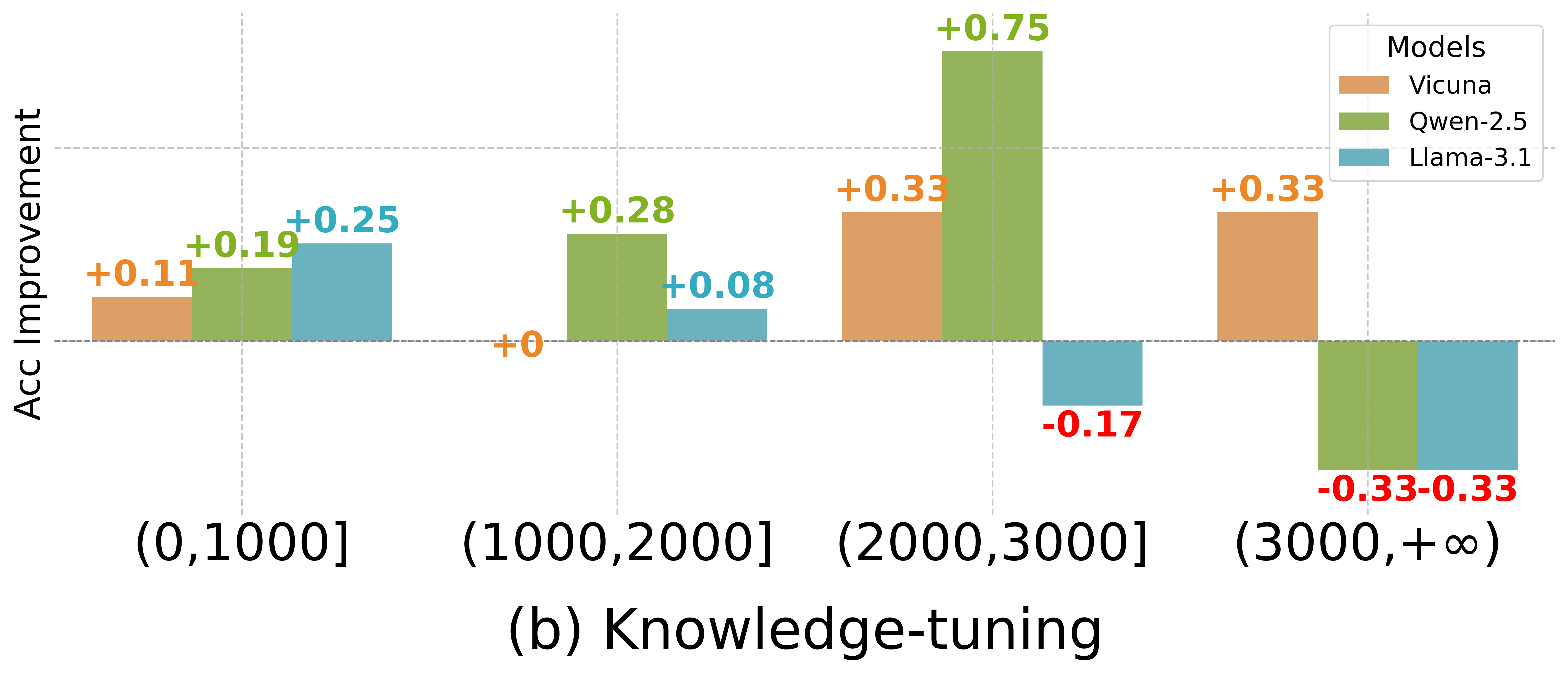}
    \end{minipage}
    \caption{\textbf{Perception-tuning and Knowledge-tuning underperform on low-prior (high Rank$_e$) entities.} The figure illustrates performance improvements compared to Zero-shot: Perception-tuning shows a significant drop for Qwen-2.5 when $Rank_e > 3000$. Similarly, Knowledge-tuning leads to notable performance declines for both Qwen-2.5 and Llama-3.1 in the low-prior range ($Rank_e > 3000$).}
    \vspace{-3mm}
    \label{fig:perspective}
\end{figure}


\textbf{Finding 2: Domain-specific finetuning with only end-to-end VQA data is insufficient}, particularly for entities with low visual prior knowledge. \cref{fig:perspective} illustrates the accuracy improvements of Perception-tuning and Knowledge-tuning models compared to Zero-shot baselines under CLIP encoder configuration. As shown in Figure (a), after Perception-tuning, Qwen-2.5 performance decreased in the $Rank_e>3000$ range, while Vicuna and Llama-3.1 showed no improvement. As shown in Figure (b), after Knowledge-tuning, Qwen-2.5 and Llama3.1's performance decreased for approximately 33\% in the $Rank_e>3000$ range compared to Zero-shot. The comprehensive experimental results across all nine encoder-language model combinations are shown in \cref{tab:method}.

\begin{figure*}[t]
  \centering
  \includegraphics[width=0.95\linewidth]{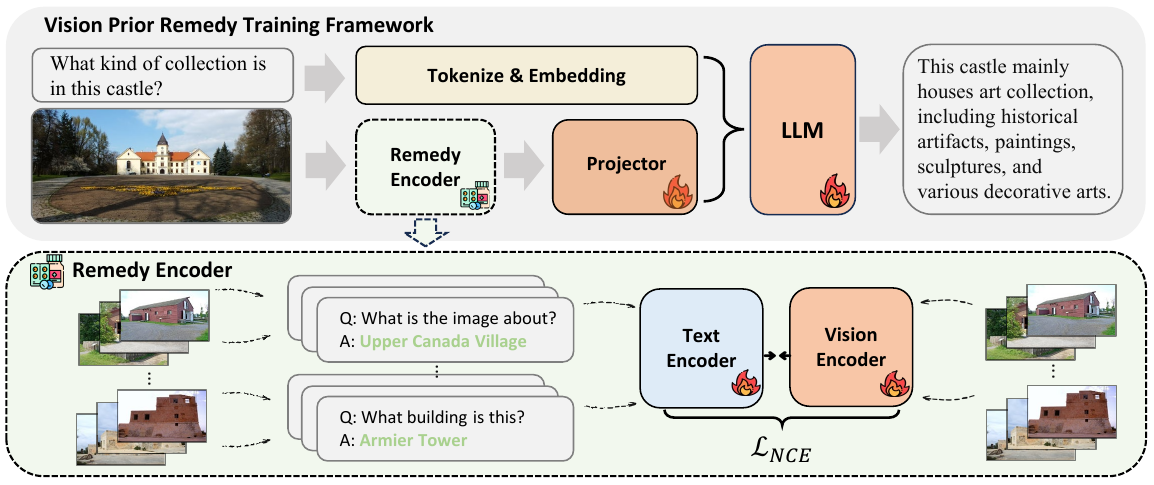}
  \vspace{-1mm}
  \caption{\textbf{Overview of our proposed VisPRE framework.} Our framework enriches the vision encoder with entity-specific prior knowledge by first extracting (image, entity\_name) pairs from Perception-tuning data and then finetuning the vision encoder using contrastive loss. The enhanced encoder is subsequently integrated into the MLLM, which is further fine-tuned on Knowledge-tuning data.}
  \label{fig:liucheng}
\end{figure*}

\subsection{Vision Prior Remediation}
\label{sec:3-4}

\begin{figure}[t]
  \centering
  \includegraphics[width=0.9\linewidth]{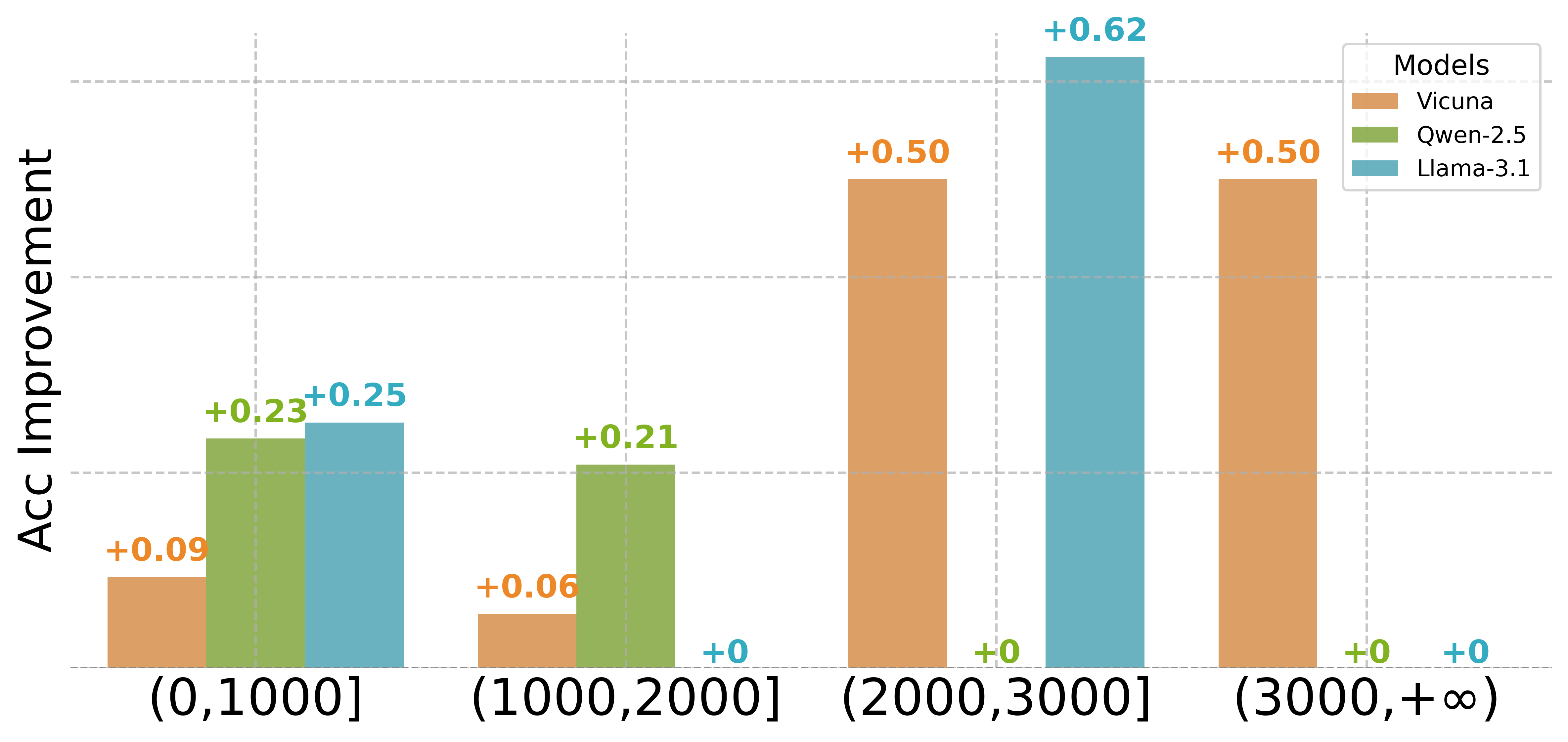}
  \vspace{-3mm}
  \caption{\textbf{VisPRE outperforms on all Rank$_e$ levels.
} The figure shows performance gains over Zero-shot: With the CLIP encoder, all three models demonstrate improvements across different $Rank_e$ entities, especially for low-prior (high $Rank_e$) entities.}
  \vspace{-3mm}
  \label{fig:ourzero}
\end{figure}



In previous sections, we established that MLLM performance correlates positively with vision prior knowledge, and that end-to-end fine-tuning yields insufficient. Based on these findings, we propose VisPRE, a training framework that injects entity-related prior knowledge at the vision encoder level to enhance MLLM performance. The specific process of our training framework is illustrated in \cref{fig:liucheng}, which comprises two key stages:

\begin{table*}[t]
\centering
\resizebox{0.8\textwidth}{!}{%

\begin{tabular}{cccccccc}
    \toprule
    Vision Encoder & LLM   & Zero-shot & Perception & Knowledge & Knowledge* & Mix*  & VisPRE(Ours) \\
    \midrule
    \multirow{3}[2]{*}{OpenAI ViT-L-14} & Vicuna-7B & 51.22  & 49.91  & 54.05  & 53.48  & 55.37  & \textbf{56.31 } \\
          & Llama3.1-8B & 37.82  & 39.26  & 45.67  & 45.99  & 44.71  & \textbf{48.24 } \\
          & Qwen2.5-7B & 46.05  & 48.84  & 54.57  & \textbf{56.59 } & 53.49  & 54.42  \\
    \midrule
    \multirow{3}[2]{*}{SigLIP ViT-SO-14 } & Vicuna-7B & 52.03  & 53.66  & 53.66  & 57.24  & 57.07  & \textbf{57.89 } \\
          & Llama3.1-8B & 38.91  & 37.66  & 41.28  & \textbf{41.84} & 41.42  & 41.28  \\
          & Qwen2.5-7B & 36.45  & 36.31  & 41.13  & 41.42  & 42.84  & \textbf{44.54 } \\
    \midrule
    \multirow{3}[2]{*}{DFN ViT-H-14} & Vicuna-7B & 59.07  & 58.70  & 63.33  & 64.97  & 62.90  & \textbf{66.85 } \\
          & Llama3.1-8B & 38.70  & 39.84  & 45.08  & 46.99  & 45.69  & \textbf{48.29 } \\
          & Qwen2.5-7B & 40.45  & 38.10  & 43.33  & 44.66  & \textbf{46.76} & 43.69  \\
    \bottomrule
\end{tabular}%

}
\vspace{-1mm}
\caption{\textbf{Results on 9 MLLM combinations.} Our method outperforms finetuning approaches including Perception-tuning, Knowledge-tuning, Knowledge-tuning* and Mix-tuning*, demonstrating that our method significantly enhances MLLM performance through prior remediation. We mark the best result in \textbf{bold} for each model, and * indicates unfreezing the vision encoder parameters during fine-tuning.}

\vspace{-2mm}
\label{tab:method}
\end{table*}

\begin{itemize}[leftmargin=*]
\item \textbf{Remedy Encoder:} We first reformat the Perception-tuning data into (image, entity\_name) pairs, and then fine-tune the vision encoder alongside the text encoder using contrastive loss. This stage enhances the encoder's prior knowledge of entities present in the Perception-tuning data.
\item \textbf{Instruction Tuning:} We incorporate the fine-tuned encoder into the MLLM architecture and perform end-to-end fine-tuning of the entire model using Knowledge-tuning data. This stage aligns the trained vision encoder with the base LLM and stimulates the model’s knowledge of entities.
\end{itemize}

To systematically evaluate VisPRE, we establish several baselines: Zero-shot, Perception-tuning, and Knowledge-tuning from Section \ref{sec:3-2}. Additionally, we include Knowledge-tuning* and Mix-tuning*, where the asterisk (*) denotes unfreezing the vision encoder parameters during fine-tuning. Mix-tuning represents a combination of Knowledge-tuning and Perception-tuning data. The evaluation results are presented in \cref{tab:method}.


\textbf{Finding 3: Remediating prior knowledge at the vision encoder level is effective.} Perception-tuning shows only marginal improvements over Zero-shot performance, occasionally even degrading results. Knowledge-tuning yields limited gains, with Knowledge-tuning* showing only modest improvement over standard Knowledge-tuning. Mix* doesn't exceed Knowledge* performance. In contrast, our VisPRE framework outperforms all baselines, achieving superior results in six of nine model combinations. As shown in \cref{fig:ourzero}, \textbf{VisPRE improves MLLM performance across all Rank$_e$ entities, particularly those with low vision priors}, demonstrating clear advantages over alternative tuning approaches in \cref{fig:perspective}. These results confirm that enhancing encoder prior knowledge substantially expands MLLM capabilities.

\section{Case Study}
\label{sec:4-3}

\begin{figure*}[h]
  \centering
  \includegraphics[width=\linewidth]{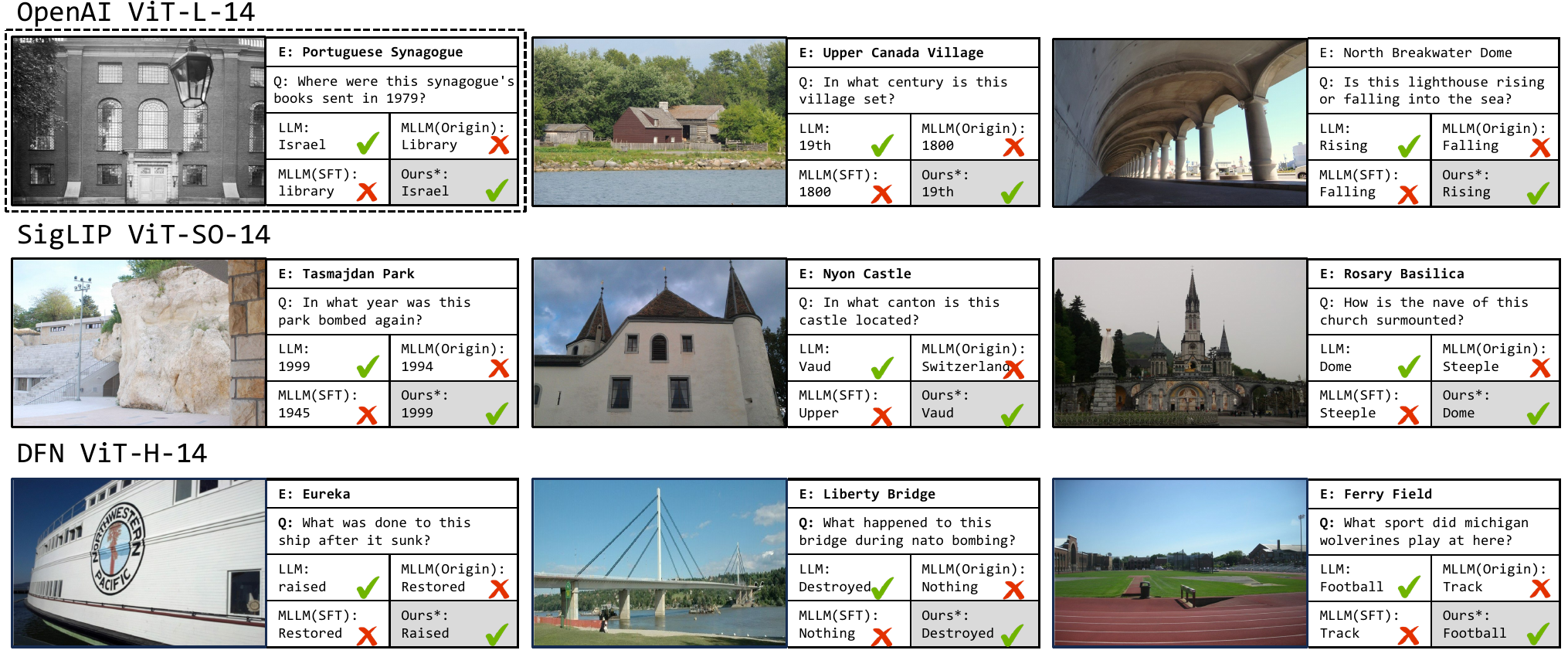}
  \vspace{-3mm}
  \caption{\textbf{Examples of Vicuna-7b's responses with different encoders.} When prompted with image description, the LLM answers correctly, demonstrating adequate knowledge of image entities. However, the original (Origin) and fine-tuning with Knowledge-tuning data (SFT) MLLM fails to answer, highlighting the limitations of its vision encoder. With VisPRE(Ours*), the model answer accuratly. For additional cases, refer to Appendix \ref{app:c}.}
  \vspace{-3mm}
  \label{fig:casestudy}
\end{figure*}

Here we present an illustrative example. As shown in the upper left of \cref{fig:casestudy}, we input an image of the Portuguese Synagogue with the entity-related question: ``Where were this synagogue's books sent in 1979?''. For \textbf{(1) LLM}: The MLLM correctly answers when receiving only the textual description ``This is Portuguese Synagogue'' instead of the actual image, indicating the LLM component possesses knowledge about this entity. For \textbf{(2) MLLM (Original)}: With image input, the MLLM fails to answer correctly. We calculated this entity's $Rank_e$ as 516, indicating low prior knowledge in the visual encoder. \textbf{(3) MLLM (SFT)}, despite end-to-end fine-tuning, still fails since the visual encoder's prior knowledge remains unchanged. Our training framework, VisPRE, first injects prior knowledge into the visual encoder, elevating the entity's $Rank_e$ to 10, then conducts end-to-end fine-tuning. Consequently, \textbf{(4) Ours*} overcomes the visual encoder's limitations and correctly answers the question.
\section{Related Works}
\label{sec:related}

\paragraph{Multi-modal Large Language Models.} MLLMs incorporate visual features into language models, enabling them to perform a wide range of visual tasks. The current MLLM implementations can be classified into two categories. (1) Monolithic MLLMs. Tokenizing different modal inputs uniformly and training the model from scratch \citep{team2024chameleon,fuyu-8b,chen2024internvl,zhan-etal-2024-anygpt}, which is computationally expensive. (2) Modular MLLMs. Utilizing pre-trained vision-language models (e.g., CLIP \cite{radford2021learning}, SigLIP \cite{zhai2023sigmoid}, DINOv2 \cite{oquab2023dinov2}) to obtain visual representations of images, and then train MLLMs through cross-modal data, aligning the visual features provided by vision encoder to language model's embedding space. This method is more data-efficient and widely used by open-source MLLMs (e.g., Flamingo \cite{alayrac2022flamingo}, BLIP2 \cite{li2023blip}, LLaVA \cite{liu2024visual}, Qwen-VL \cite{bai2023qwen}, InternVL2 \cite{chen2024internvl}). Our work focuses on modular multimodal models. While most works treat modular MLLM as a unified system, our research focuses on the impact of vision encoder part on the language model part.

\paragraph{Cross-modality Alignment.} With increasing adoption of Modular MLLMs, research focuses on the relationship between vision encoders and MLLM performance. \citet{tong2024eyes} found CLIP \cite{radford2021learning} and corresponding MLLMs have similar performance trends across visual modalities, indicating CLIP features cause MLLM deficiencies in these modes, and addressed these by introducing DINOv2 \cite{oquab2023dinov2} features. \citet{yang2024law} proposed cross-modal alignment metrics to measure vision encoder performance, fitting a binary quadratic polynomial that predicts MLLM performance using that encoder. Different from previous works, our research offers a novel perspective, demonstrating that MLLM performance correlates positively with its vision encoder's prior knowledge.

\section{Conclusion}
\label{sec:conclusion}
In this paper, we introduce $Rank_e$ to quantify prior knowledge in vision encoder. We find that MLLM's performance is positively correlated with prior knowledge of vision encoder, and end-to-end finetuning MLLM yields insufficient on improving low prior entity performance. To address this issue, we propose VisPRE training framework that enhances MLLM’s performance by increasing the prior knowledge within the vision encoder. Our study demonstrates a novel pathway for enhancing MLLM performance, offering substantial value for applications involving uncommon entities.

\section*{Limitations}

The primary limitation of our study is the current unavailability of VQA datasets with comprehensive rare entity annotations. While our study explores MLLMs' capabilities when confronted with uncommon entities—those inadequately represented in visual encoders' pretraining data, most established entity-annotated datasets like S3VQA \cite{jain2021select} predominantly feature common entities. To address this challenge, we leveraged the Encyclopedia VQA \cite{mensink2023encyclopedic} dataset with its diverse collection of 16.7k entity types, providing a sufficient foundation to identify and analyze less familiar entities. Nevertheless, our findings would benefit from additional specialized datasets explicitly focused on uncommon entities, which would enable a more granular analysis of visual encoders' boundary capabilities and offer complementary insights to our current observations.

\section*{Ethics Statement}

Our study utilizes MLLMs for knowledge-based VQA tasks. MLLMs may reflect biases present in the training data. Additionally, the VQA data used in our research includes pictures of landscapes and related knowledge questions, which may lead the model to generate offensive content. In this regard, we suggest users to examine the generated outputs cautiously in real-world applications.

\bibliography{custom}

\clearpage


\appendix

\section{Datasets}
\label{app:a}

Here we describe the detailed construction process of our dataset. Based on Encyclopedia-VQA \cite{mensink2023encyclopedic}, we constructed Knowledge-tuning and Perception-tuning datasets for each encoder-language model combination to validate \textbf{Finding 2}.

\subsection{Preprocess}
\label{sec:A-1}

\textbf{Question Filtering.} First, we focus on improving the parts where MLLM's capabilities are limited by the vision encoder. Therefore, we only retained questions that could be answered by the corresponding LLM when prompted with ``This is \{entity\_name\}'' instead of the actual image. Next, to ensure that there were no duplicate or similar questions for the same entity across training and test sets, we deduplicated the dataset based on (entity\_name, answer) pairs. Finally, we only retained entities with three or more corresponding questions to ensure sufficient questions for dividing into training and validation sets.

\begin{figure*}[t]
  \centering
    \begin{minipage}[b]{\textwidth}
        \centering
        \includegraphics[width=\textwidth]{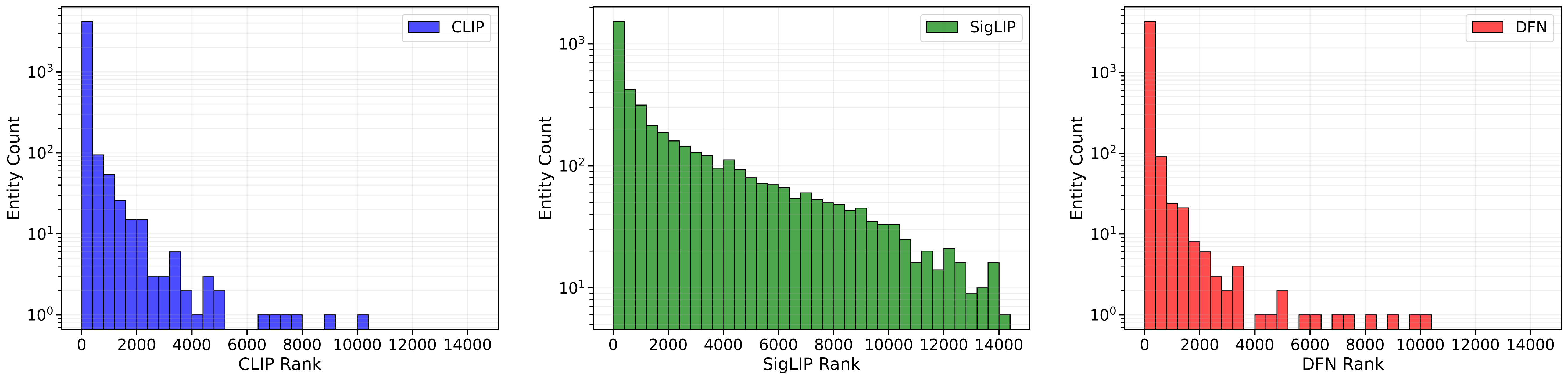}
        (a) Vicuna
    \end{minipage}
    \begin{minipage}[b]{\textwidth}
        \centering
        \includegraphics[width=\textwidth]{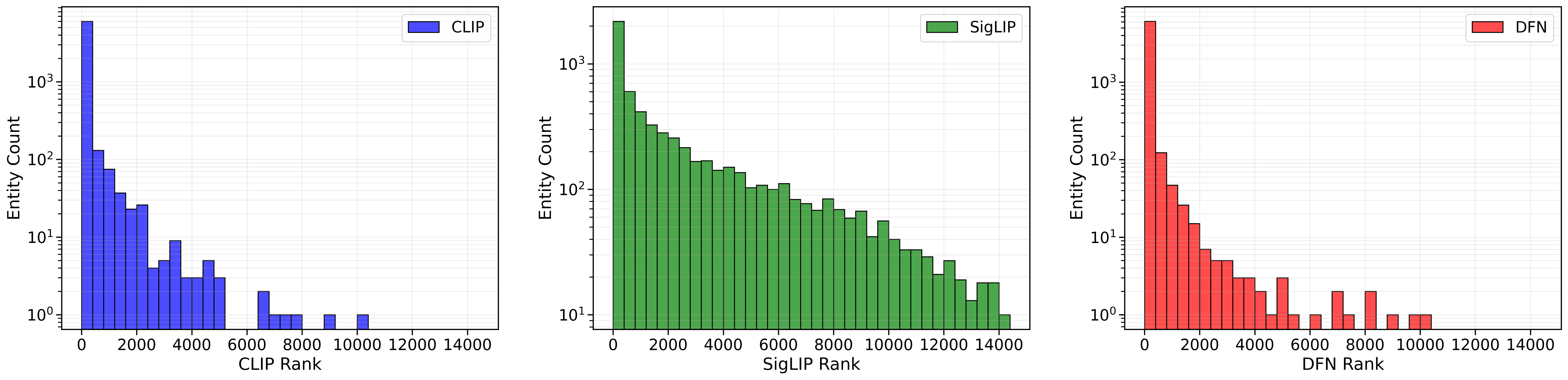}
        (b) Qwen-2.5
    \end{minipage}
    \begin{minipage}[b]{\textwidth}
        \centering
        \includegraphics[width=\textwidth]{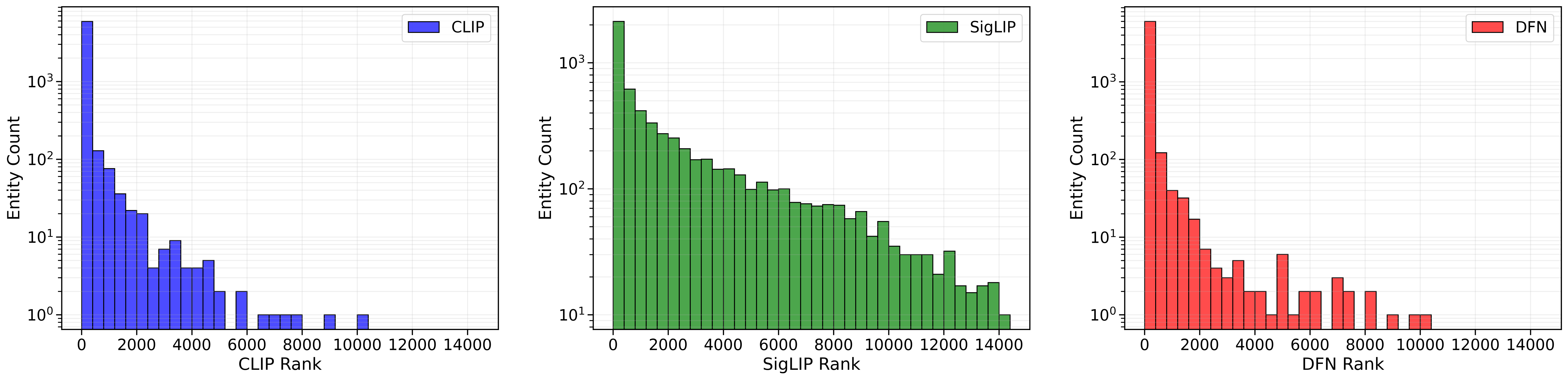}
        (c) Llama-3.1
    \end{minipage}

  \caption{The $Rank_e$ distribution of entities calculated using three different encoders. Here we show the entities that (a)Vicuna, (b)Qwen-2.5 and (c)Llama-3.1 could answer after using text prompts instead of entity images. We can see that the $Rank_e$ distributions for both CLIP and DFN are concentrated in intervals near the left side, while SigLIP's $Rank_e$ distribution is relatively uniform.}
  \label{fig:rank_distribution}
\end{figure*}

\noindent\textbf{Prior Calculation.} We calculated $Rank_e$ for all entities in the filtered dataset. We examined the distribution of $Rank_e$ calculated using different types of encoders (CLIP \cite{radford2021learning}, SigLIP \cite{zhai2023sigmoid}, DFN \cite{fang2023data}) across the dataset, as shown in \cref{fig:rank_distribution}. We found significant variations in $Rank_e$ distributions among different encoders. CLIP's $Rank_e$ values were mostly concentrated in the range of $Rank_e<400$, with entity counts increasing as $Rank_e$ decreased; In contrast, SigLIP's $Rank_e$ distribution is more uniform, with at least 10 entities present across most $Rank_e$ intervals; DFN's $Rank_e$ distribution was similar to CLIP's, with most values concentrated in the range of $Rank_e<400$.

\noindent\textbf{Entity Sampling.} For SigLIP, we divided $Rank_e$ into intervals of size 1000 and sampled 10 entities from each interval. For CLIP and DFN, using the same sampling strategy as SigLIP would result in insufficient sampling of entities in dense intervals, making it difficult to distinguish different levels of prior knowledge in these regions. Therefore, we adopted a sampling method that approximates the original distributions of CLIP and DFN. We sampled 10 entities from intervals of $0<Rank_e<=2$, $2<Rank_e<=4$, $4<Rank_e<=8$, ..., $512<Rank_e<=1024$, $Rank_e>1024$, ensuring that the sampled distribution approximates the original distribution while retaining all entities with low prior knowledge to reflect the relationship between entity prior knowledge and model performance. Finally, we retained the questions corresponding to the sampled entities and divided the dataset into training and test sets, with statistical information shown in \cref{tab:data_statis}.

\begin{figure}[t]
  \centering
  \includegraphics[width=\linewidth]{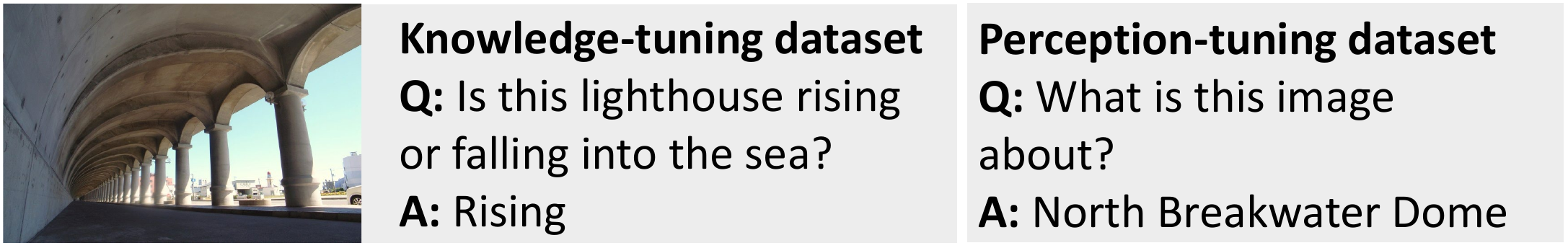}
  \vspace{-6mm}
  \caption{Knowledge-tuning and Perception-tuning datasets}
  \vspace{-6mm}
  \label{fig:cvqa_example}
\end{figure}

\subsection{Construction}

For Knowledge-tuning dataset, we use the original question and answer from the Encyclopedia-VQA dataset. For Perception-tuning dataset, we replace the original question in the Knowledge-tuning dataset with cognitive question like ``What is this image about?'' and substitute the answers with the entity text corresponding to the image. Examples of Knowledge-tuning and Perception-tuning datasets are shown in \cref{fig:cvqa_example}.

\section{Evaluation Settings}
\label{app:b}
We employ Llama-3.1-70B \cite{dubey2024llama} to evaluate the accuracy of MLLM's responses to VQA questions. Specifically, we provide Llama-3.1-70B with the question, entity name (\texttt{wikipedia\_title} in prompt), ground truth answer, and MLLM's response. The model outputs \textit{true} to indicate a correct answer and \textit{false} to indicate an incorrect answer. The prompt template is shown in \cref{fig:evaluation_prompt}, with the \texttt{few\_shot\_examples} shown in \cref{fig:few_shot_examples}.

\section{More Cases}
\label{app:c}
\begin{figure*}[h]
  \centering
  \includegraphics[width=\linewidth]{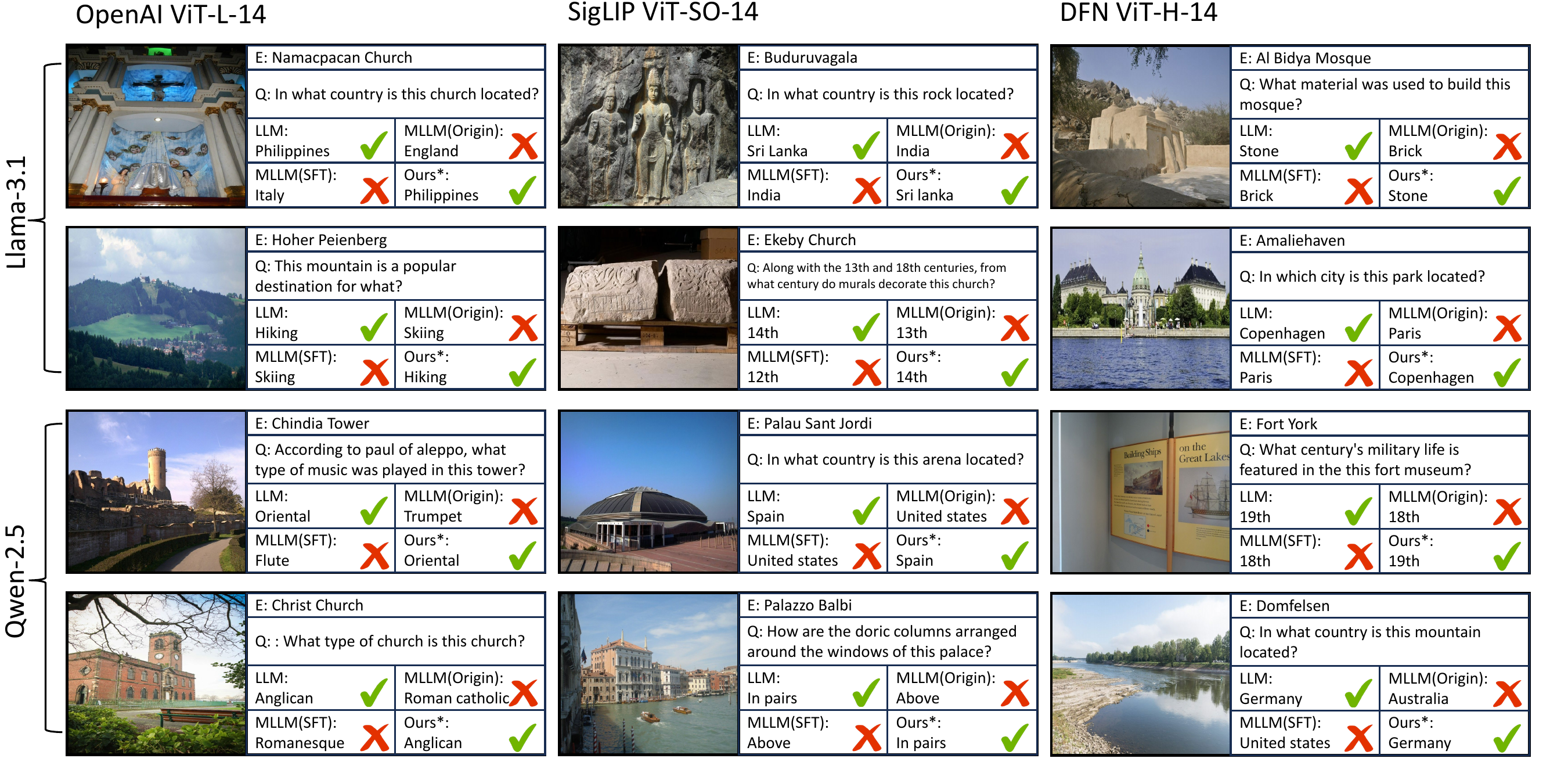}
  \caption{We present examples of Llama-3.1 and Qwen-2.5's responses under three encoder setups. When prompted with text to identify objects in the image, the LLM provides correct answers, demonstrating its knowledge of image entities. In contrast, the MLLM (Origin) fails to respond correctly, highlighting the limitations of its vision encoder. Even after fine-tuning with Knowledge-type VQA data (MLLM SFT), the model still cannot provide accurate answers, revealing the constraints of fine-tuning. Finally, with our Remedy Encoder, the model delivers accurate responses, demonstrating that our method effectively expands the MLLM's visual priors.}
  \label{fig:casestudy2}
\end{figure*}

In \cref{fig:casestudy2}, we demonstrated Vicuna-7B's responses under different encoder configurations. Here in \cref{fig:casestudy2}, we show examples of responses from Llama-3.1-7B and Qwen-2.5-7B under different encoders.

\begin{figure*}[t]
\begin{tcolorbox}[colframe=cyan!40!black, title=\textbf{Prompt for Llama-3.1 evaluation}]

You are an expert evaluator tasked with assessing the correctness of model predictions. Your job is to determine if a given prediction is correct based on the provided information. Follow these strict guidelines:\\

1. You will be given four pieces of information:\\
\ \ \ - Question: The original question asked\\
\ \ \ - Wikipedia\_title: The title of the Wikipedia article that corresponds to the knowledge base for the question\\
\ \ \ - Answer: The correct answer(s) to the question, possibly including multiple candidates separated by "$\vert$"\\
\ \ \ - Prediction: The model's prediction to be evaluated\\

2. Understand that the question is specifically about the entity described in the Wikipedia\_title.\\

3. Compare the prediction to the answer(s), taking into account the context of the question and the Wikipedia\_title.\\

4. Apply these strict criteria:\\
\ \ \ - The prediction must be accurate and specific.\\
\ \ \ - If there are multiple candidate answers separated by "$\vert$", the prediction must match at least one of them to be considered true.\\
\ \ \ - For numerical answers, the prediction must be within 10\% of at least one correct answer to be considered true.\\
\ \ \ - For categorical or descriptive answers, the prediction must match the key concepts or categories in at least one of the provided answers.\\
\ \ \ - Partial or vague answers that don't fully capture the specificity of any correct answer should be considered false.\\
\ \ \ - Pay close attention to units, specificity, and context provided in the question, Wikipedia\_title, and answer(s).\\

5. Your response must be exactly one word:\\
\ \ \ - Output "true" if the prediction meets all the criteria for correctness.\\
\ \ \ - Output "false" if the prediction fails to meet any of the criteria.\\

6. Do not provide any explanations or additional comments.\\

\{\texttt{few\_shot\_examples}\}\\

Remember, your task is to evaluate the correctness of the prediction based on all the information provided. Be strict in your assessment, but consider all given correct answers. Respond only with "true" or "false".\\

Question: \{\texttt{question}\}\\
Wikipedia\_title: \{\texttt{wikipedia\_title}\}\\
Answer: \{\texttt{answer}\}\\
Prediction: \{\texttt{prediction}\}\\
Evaluation: \\

\end{tcolorbox}
\vspace{-4mm}
\caption{Complete prompt for evaluating MLLM responses using Llama-3.1-70B. We prompt the model to determine whether a \texttt{prediction} is correct by examining the \texttt{question}, \texttt{wikipedia\_title} (entity name), and \texttt{answer}. The model outputs \textit{true} for correct predictions and \textit{false} for incorrect ones. The \texttt{few\_shot\_examples} are shown in \cref{fig:few_shot_examples}}
\label{fig:evaluation_prompt}
\end{figure*}
\begin{figure*}[t]
\begin{tcolorbox}[colframe=cyan!40!black, title=\textbf{Few-shot examples}]

Examples:\\
Question: Along with the mojave desert, in what desert is this plant found?\\
Wikipedia\_title: Acmispon rigidus\\
Answer: Sonoran Desert\\
Prediction: Sonoran\\
Evaluation: true\\

Question: How many people can this stadium host?\\
Wikipedia\_title: Mercedes-Benz Stadium\\
Answer: 71,000 $\vert$ 75,000\\
Prediction: 73,000\\
Evaluation: true\\

Question: When was this novel first published?\\
Wikipedia\_title: To Kill a Mockingbird\\
Answer: 1960\\
Prediction: 1962\\
Evaluation: false

\end{tcolorbox}
\caption{\texttt{few\_shot\_examples} in prompt for Llama-3.1 evaluation. We provide three examples to help the model understand the evaluation requirements.}
\label{fig:few_shot_examples}
\end{figure*}

\end{document}